\documentclass[sigconf]{acmart}
\usepackage{algorithm2e}

\AtBeginDocument{%
  \providecommand\BibTeX{{%
    \normalfont B\kern-0.5em{\scshape i\kern-0.25em b}\kern-0.8em\TeX}}}

\acmConference[]{}{}{}

\begin{document}

\title{An Adversarial Approach for the Robust Classification of Pneumonia from Chest Radiographs}


\author{Joseph D. Janizek}
\email{jjanizek@cs.washington.edu}
\affiliation{%
  \institution{Paul G. Allen School of Computer Science and Engineering}
}

\author{Gabriel Erion}
\email{erion@cs.washington.edu}
\affiliation{%
  \institution{Paul G. Allen School of Computer Science and Engineering}
}

\author{Alex J. DeGrave}
\email{degrave@cs.washington.edu}
\affiliation{%
  \institution{Paul G. Allen School of Computer Science and Engineering}
}
\author{Su-In Lee}
\email{suinlee@cs.washington.edu}
\affiliation{%
  \institution{Paul G. Allen School of Computer Science and Engineering}
}

\renewcommand{\shortauthors}{Janizek et al.}
\newcommand{\sam}{\textcolor{violet}}
\newcommand{\joe}{\textcolor{red}}

\begin{abstract}
  While deep learning has shown promise in the domain of disease classification from medical images, models based on state-of-the-art convolutional neural network architectures often exhibit performance loss due to dataset shift. Models trained using data from one hospital system achieve high predictive performance when tested on data from the same hospital, but perform significantly worse when they are tested in different hospital systems. Furthermore, even within a given hospital system, deep learning models have been shown to depend on hospital- and patient-level confounders rather than meaningful pathology to make classifications. In order for these models to be safely deployed, we would like to ensure that they do not use confounding variables to make their classification, and that they will work well even when tested on images from hospitals that were not included in the training data. We attempt to address this problem in the context of pneumonia classification from chest radiographs. We propose an approach based on adversarial optimization, which allows us to learn more robust models that do not depend on confounders. Specifically, we demonstrate improved out-of-hospital generalization performance of a pneumonia classifier by training a model that is invariant to the view position of chest radiographs (anterior-posterior vs. posterior-anterior). Our approach leads to better predictive performance on external hospital data than both a standard baseline and previously proposed methods to handle confounding, and also suggests a method for identifying models that may rely on confounders.
\end{abstract}

\maketitle

\section{Introduction}

A variety of recent papers have demonstrated the promise of deep learning for medical imaging tasks. From the prediction of diabetic retinopathy using retinal scan images to the diagnosis of melanoma from photographs, machine learning approaches have achieved near-physician level performance \cite{Gulshan2016DevelopmentPhotographs,Esteva2017Dermatologist-levelNetworks}. Deep learning classifiers of chest radiographs are not only promising in a research setting, but have also been deployed in clinical practice. For example, an algorithm to detect 4 different thoracic diseases from frontal chest radiographs was evaluated in an emergency medicine setting and was found to increase radiology residents' sensitivity \cite{Hwang2019DeepDepartment}.

Despite these major advances, there are still significant limitations for medical deep learning. One of these problems is dataset shift, or the loss in performance when a model is tested on data that is drawn from a different distribution than the data used for training the model \cite{Quionero-Candela2009DatasetLearning,Pooch2019CanClassification}. Zech et al. \cite{Zech2018} found that a deep learning pneumonia classifier trained on data from two hospital systems exploited differences in the base rate of pneumonia between the two hospitals by learning to identify each radiograph's hospital of origin rather than anatomically-relevant features of pneumonia. While this model apparently had high predictive performance, when the model was tested on radiographs from a third hospital not present in the training data its performance significantly decreased. Furthermore, even within a single hospital system, confounded predictions may be a problem for deep learning. For example, Badgeley et al. \cite{Badgeley2019DeepVariables} demonstrated that a deep learning hip fracture classifier was leveraging patient-level variables (such as age and gender) and process-level variables (such as scanner model and hospital department) in its predictions. After controlling for these variables during model evaluation by rebalancing the test set, they found that the classifier performed no better than random. A recent multi-society statement on the ``Ethics of Artificial Intelligence in Radiology'' points to the importance of being able to understand and guide the decision-making process of machine learning algorithms to ensure that these algorithms can be safely and effectively used in clinical practice \cite{Geis2019EthicsStatement}. While the works  above have described the brittleness of deep learning medical imaging classifiers, more work is needed to create robust models.

We propose an approach based on adversarial neural networks to address dataset shift by learning models that are invariant to confounders that may shift across hospitals. In particular, we focus on the problem of pneumonia classification from chest radiographs, as the problem of confounding and dataset shift has been particularly well-documented for this task \cite{Zech2018}. We find that (1) potential model confounding can be effectively identified by evaluating how well confounders can be predicted from a model's output, that (2) adversarial training enables pneumonia classification that is independent of radiograph view, and that (3) the adversarially-trained models attain better generalization performance when tested in novel hospital systems. \footnote{Code to reproduce this project is available at https://github.com/suinleelab/cxr\_adv}

\section{Problem Statement}
\label{sec:ProbStatement}

\begin{figure}
\centering
\includegraphics[width=0.65\linewidth]{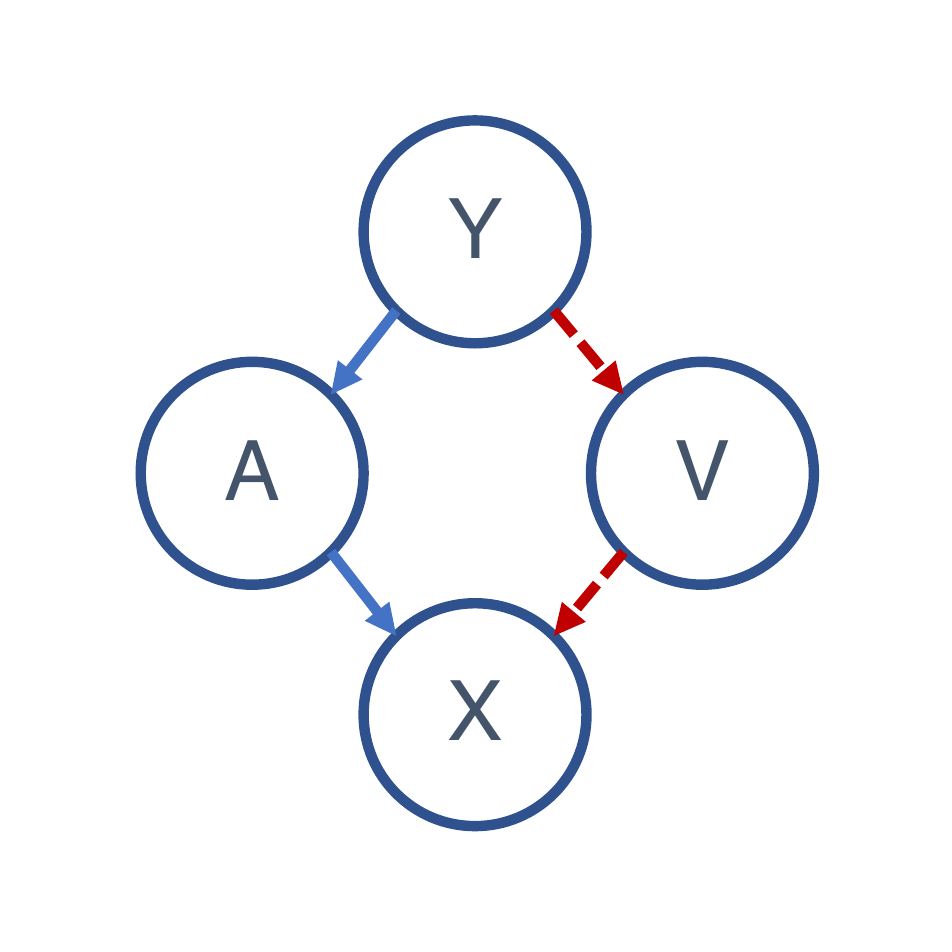}

\caption{Causal graph showing relationships that form part of one plausible data generating process for chest radiographs: relationships are between pneumonia (Y), radiograph view position (V), anatomically relevant radiographic features (A), and the final chest radiograph (X). Red and dashed edges indicate a view-mediated causal path between the radiograph and pneumonia that may shift between different datasets or hospitals. We emphasize that this does not illustrate the full data generating process, and that many data generating processes are possible.}
\label{fig:CausalGraph}
\end{figure}


We first consider some of the causal relationships forming part of one plausible data generating process for chest radiographs, given by the random variable $X$ in \autoref{fig:CausalGraph}. A patient's pneumonia status, given by the random variable $Y$, will lead to a variety of anatomically-relevant features $A$, such as increased radiopacity or consolidation in the lung fields, that form part of the radiograph. Furthermore, the patient's disease status will lead to a variety of clinical signs and symptoms which will influence which department they are seen in (e.g. in-patient or out-patient). Different departments may use different scanners (portable or fixed) and these scanners may be taken with different views ($V$). Frontal chest radiographs may be taken with either an anterior-posterior (AP) view where the x-ray source is positioned such that x-rays enter through the front of the chest and exit through the back of the chest, or a posterior-anterior (PA) view where the x-ray source is positioned such that x-rays enter through the back of the chest and exit through the front.

View directly impacts the appearance of chest radiographs in a variety of ways. Different views cause anatomical structures to have different relative sizes in radiographs since their distance from the radiographic source is altered \cite{Puddy2007InterpretationRadiograph}. Furthermore, AP radiographs are taken on portable scanners, which may place text such as ``PORTABLE'' or ``SEMI-UPRIGHT'' directly on the image. For this graph, it is plausible that the relationship between pneumonia and view may not be consistent across hospitals. The AP view position is generally associated with a higher prevalence of disease, as sicker patients are more likely to need to have a portable scanner brought to them \cite{Puddy2007InterpretationRadiograph,Zech2018WhatLearning}. In our source training dataset (described below in \autoref{sec:Data}), however, the standard relationship is reversed and the prevalence of pneumonia is 2-fold \emph{higher} in PA view radiographs ($2.1$\% base rate of pneumonia in AP images vs. $3.9$\% base rate of pneumonia in PA images). As the difference in base rate between the subgroups increases, the worse the generalization performance should be (see \autoref{sec:BaseRateAlteration}). Since the relationships between pneumonia and view may not be consistent across hospitals, we hypothesize that by learning a model that is invariant to differences in radiograph view, we can create a model that will be more robust to dataset shift. View is additionally an important confounder to control because commerically-available chest radiograph algorithms are currently designed to accept \emph{both} AP and PA view radiographs as input \cite{Hwang2019DeepDepartment}.

We formally state the problem as follows. We are given data from a source distribution $\mathcal{S}$ where each sample (indexed by $i$) is a 3-tuple consisting of a radiograph $x_i \sim X$, a multi-label classification label $y_i \sim Y$, and a binary indicator of view $v_i \sim V$. We would like to learn a model that outputs a pneumonia score that will generalize well to a target domain $\mathcal{T}$, where the relationship between the nuisance variable and the outcome may be different in the target domain than in the source domain. In our particular problem, we assume that we have no access at all to data from the target distribution, corresponding to what Subbaswamy et al. \cite{Subbaswamy2019FromAI} refer to as a proactive approach to addressing dataset shift. Much of the prior work on adversarial domain adaptation has corresponded to a different problem, in which we assume that we have access to \emph{unlabeled data} from the target distribution, corresponding to what Subbaswamy et al. \cite{Subbaswamy2019FromAI} refer to as a reactive approach to dataset shift \cite{Ben-David2010ADomains,Ganin2016Domain-adversarialNetworks,Mahmood2018UnsupervisedTraining,Javanmardi2018DomainTraining}. Since we have no data from the target distribution, we instead aim to learn a classifier $f$ that outputs a pneumonia score $S$ such that $S \perp V$. Even though we use all 13 of the different pathologies in $Y$ to train our model, since we only require that our model learns a relationship such that $S \perp V$, and not $Y \perp V$, there is no constraint for the model to learn view-independent scores for any of the other non-pneumonia pathologies.

\section{Methods}

\subsection{Data}
\label{sec:Data}

To assess the robustness of models to dataset shift, we used chest radiographs from two large publicly-available datasets. For our model training source domain, we used the CheXpert dataset from Stanford \cite{Irvin2019Chexpert:Comparison}. This dataset contains 224,316 chest radiographs of 65,240 patients. We considered only the 191,229 frontal radiographs (AP or PA view) in the dataset, excluding all of the lateral radiographs. Since the test split in the original CheXpert dataset only contained 8 radiographs that were positive for pneumonia, all of which were AP radiographs, we moved 92 more positive pneumonia radiographs (for a total of 100 positive pneumonia radiographs) to the test set for the sake of better pneumonia performance evaluation. For our target domain, we used the MIMIC-CXR dataset from Massachusetts Institute of Technology \cite{Johnson2019MIMIC-CXR:Radiographs}. This dataset includes 371,920 chest radiographs of 65,079 patients. After filtering lateral radiographs, we had 249,995 frontal radiographs remaining. One major advantage of using these two datasets is that they have the same set of 13 labels (``Enlarged Cardiomediastinum,'' ``Cardiomegaly,'' ``Lung Opacity,'' ``Lung Lesion,'' ``Edema,'' ``Consolidation,'' ``Pneumonia,'' ``Atelectasis,'' ``Pneumothorax,'' ``Pleural Effusion,'' ``Pleural Other,'' ``Fracture,'' and ``Support Devices'') and are created using the same labeling algorithm. This algorithm takes expert-generated free-text radiological reports associated with each chest radiograph as input and outputs the set of pathology labels. Using data labeled with the same natural language processing algorithm helps to remove the potential effects of dataset shift due to differences in the label generating process.

\subsection{Standard training}
\label{sec:StandardTraining}


To train our baseline models for prediction, we used the architecture and training procedure described in \cite{Zech2018} and \cite{Rajpurkar2017CheXNet:Learning}. The model architecture used was a DenseNet-121 initialized with weights pretrained on ImageNet, which can be downloaded from the PyTorch torchvision models subpackage \cite{Huang2016DenselyNetworks,Paszke2017AutomaticPytorch}. While we were primarily interested in pneumonia detection, we found that using all pathology labels available in the CheXpert dataset during training significantly increased pneumonia classification performance. Since the number of classes in the CheXpert dataset is different than the number of classes in the ImageNet dataset, the classification head for the pretrained DenseNet-121 was removed and replaced by a linear layer with output dimensions equal to the number of labels in the CheXpert dataset, followed by a sigmoid activation function. A binary cross-entropy loss was optimized using an SGD optimizer with momentum of $0.9$, weight decay of $10^{-4}$, and an initial learning rate of $10^{-2}$. Early stopping was implemented by monitoring binary cross-entropy loss on a held out split of validation data. Our validation set, representing $5\%$ of the training data, was split on patients rather than radiograph index. If validation loss did not improve over an epoch, the learning rate was decreased by a factor of $10$. If validation loss failed to improve for $3$ consecutive epochs, training was stopped. Performance was then evaluated on the held out test set. This procedure was repeated three separate times to attain standard deviations of performance.

\subsection{Adversarial deconfounding}
\label{sec:AdvTraining}

To learn more robust models that generalize better to external test data, we propose an approach based on adversarial training. This approach consists of jointly training two neural networks. The first is the classifier, $f$, which is trained to predict a pneumonia label $y$ from a chest radiograph $x$. The second is an adversary, $g$, which is trained to predict the view $v$ from the output score $s$ of the classifier $f$. The optimization procedure consists of alternating between training the adversary network until it is optimal, then training the classifier to fool the adversary while still predicting pneumonia well.

This approach aims to proactively mitigate the potential effects of domain shift by controlling for known confounders in medical images using adversarial training. In addition to the applications for reactive domain adaptation mentioned above in \autoref{sec:ProbStatement}, adversarial training has been used in a variety of other areas to learn models or representations that are independent of a given variable. For example, there is a significant body of literature in the area of algorithmic fairness where adversarial training has been used to learn representations that are fair with respect to protected classes such as race or gender \cite{Edwards2015CensoringAdversary,Madras2018LearningRepresentations,Wadsworth2018AchievingPrediction}. In the physical sciences, adversarial training has been used to learn classifiers capable of detecting interesting particle jets in particle colliders that are independent of the presence of nuisance interactions in the collider \cite{Louppe2017LearningNetworks}.

We emphasize that one major contribution of our work compared to prior work on deep learning for medical images is that we take advantage of causal domain knowledge to improve generalization performance without needing to use \emph{any} data from the target domain. Where previous approaches to domain adaptation use adversarial training to either learn a score or intermediate representation that are \emph{domain-invariant} by augmenting training with unlabeled data from the target domain, we instead use our domain knowledge about the causal relationships involved in our data to find nuisance variables that potentially will have a different relationship with the outcome in the target domain than in the source domain. We then use an adversarial approach to learn a classifier that is invariant to the nuisance variable, which requires no data whatsoever from the target domain. 

To implement our training, we take the approach suggested in Louppe et al. \cite{Louppe2017LearningNetworks} and adapt it for use in the application of radiograph classification. For the notation in the following sections, the parameterization of classifier $f$ will be given as $\theta_f$, while the parameterization of adversary $g$ will be given by $\theta_g$. The classifier's output score for pneumonia is given by $s = f(x)^{pneumo}$ (where the $(pneumo)$ superscript indicates the index for pneumonia in the multi-label output vector).

\subsubsection{Separately pretraining classifier and adversary}

The classifier $f$ is first trained using the procedure described in the standard training section above to optimize the negative log-likelihood of $Y | X$ under $\theta_f$:

\begin{equation}
    \mathcal{L}_{f}(\theta_f) = \mathbb{E}_{x\sim X}\mathbb{E}_{y\sim Y|x} [- \log p_{\theta_f} (y | x) ].
\end{equation}

Then, the parameters of the classifier are fixed and the adversary network is trained. The architecture used for the adversary is a simple feed-forward network with $3$ hidden layers of $32$ nodes. We used ReLU activation functions between the hidden layers, and a linear output. This architecture was selected to have sufficient capacity to model non-linear dependency between the score and view while still being lightweight enough for quick optimization. The network is optimized to minimize the following objective:

\begin{equation}
    \mathcal{L}_{r}(\theta_f, \theta_r) = \mathbb{E}_{s\sim f(X; \theta_f)}\mathbb{E}_{v\sim V|s} [- \log p_{\theta_r} (v | s) ].
\label{eq:adv_loss}
\end{equation}

This means that the adversary takes the scalar-valued pneumonia score output by the classifier as its input, and outputs a scalar-valued prediction of view. The adversary was pretrained for a single epoch.

\subsubsection{Joint adversarial optimization}

After both the classifier and the adversary were pretrained, we began joint adversarial optimization. Each ``joint optimization epoch'' consisted of first fixing the classifier, then training the adversary for one epoch by minimizing the loss of the batch stochastic gradients for each of $K = N / M$ minibatches present in the entire dataset (where $N$ is the number of total samples in the training data and $M$ is the size of the minibatch):

\begin{equation}
\nabla_{\theta_r} \sum_{k=1}^{K} \sum_{m=1}^{M_k} - \log p_{\theta_r} (v_m | s_m).
\end{equation}

Then, after the adversary is trained to optimally predict the nuisance variable $V$ from the score output by the classifier, the parameters of the adversarial network $\theta_r$ are fixed, and we draw a single minibatch of data and update the model by descending the stochastic gradients of the minibatch

\begin{equation}
\nabla_{\theta_f} \sum_{m=1}^{M} \big[ -\log p_{\theta_f} (y_m | x_m) + \log p_{\theta_r} (v_m | s_m) \big].
\end{equation}

The procedure of an entire epoch of training for the adversary with the classifier fixed, and a single minibatch of training for the classifier with the adversary fixed, is repeated until the model achieves optimal performance while its output is independent of the nuisance variable. 

Louppe et al. \cite{Louppe2017LearningNetworks} showed that the optimal solution of this minimax optimization scheme is a classifier $f$ that is optimal with respect to the training data with output $S$ that is independent of $V$. If no such classifier exists, then the weight of the adversarial loss term given in \autoref{eq:adv_loss} can be tuned with an additional hyperparameter $\lambda$ to make a tradeoff between stability (in terms of independence of the classifier from the nuisance variable) and accuracy (in terms of classification performance given the data). For all of our models, we used a value of $\lambda = 1$. Finally, while other approaches have enforced independence between $V$ and some intermediate layer of the network, if we want a pneumonia score $S$ that is independent of $V$, we observe that it suffices to directly adversarially optimize the prediction of $V$ from $S$.

\subsection{Previous approaches for controlling confounders}

Attempting to control for confounding in machine learning models is a well studied problem, and has previously been specifically studied in the domain of medical imaging \cite{Rao2017PredictiveConfounds}. In addition to testing the performance of our adversarial approach, we also compared to a variety of previously used approaches for modeling medical images in the presence of confounders.

\subsubsection{Instance sampling} One approach to domain adaptation involves re-weighting samples in the training data \cite{Pearl2011TransportabilityApproach,Little2019CausalBootstrapping,Shimodaira2000ImprovingFunction,Sugiyama2008DirectAdaptation}. We re-implement the approach suggested in Rao et al., called Instance Weighting \cite{Rao2017PredictiveConfounds}. In a normal empirical risk minimization framework, we assume that the data the model will be evaluated on will be drawn from the same data generating process as that which the model is trained on, and thus aim to minimize the empirical risk:

\begin{equation}
  f^* = \textrm{argmin}_{f\in\mathcal{F}} \sum_{i=1}^{n} \frac{1}{n} \ell (f(X_i), Y_i).
\end{equation}

If we assume that we will have test data drawn from a different distribution, we can try to reweight the samples in our training set to minimize the empirical risk in the \emph{target population} instead of the source population:

\begin{equation}
f^* = \textrm{argmin}_{f\in\mathcal{F}} \sum_{i=1}^{n} \frac{1}{n} \bigg[\frac{\hat{P}^{\mathcal{T}}(V_i,Y_i)}{\hat{P}^{\mathcal{S}}(V_i,Y_i)}\bigg] \ell(f(X_i),y_i),
\end{equation}
where $\hat{P}^{\mathcal{S}}(V_i,Y_i)$ and $\hat{P}^{\mathcal{T}}(V_i,Y_i)$ indicate the joint density of radiograph view and pneumonia in the source and target domains respectively.

Since we do not have any information about the target distribution, we assume the target marginal distributions of the targets and the confounders are identical to the source marginal distributions, which means that the loss function factorizes to the following form:

\begin{equation}
f^* = \textrm{argmin}_{f\in\mathcal{F}} \sum_{i=1}^{n} \frac{1}{n} \bigg[\frac{\hat{P}^{\mathcal{S}}(Y_i)}{\hat{P}^{\mathcal{S}}(Y_i|V_i)}\bigg] \ell(f(X_i),y_i).
\end{equation}

In order to avoid particular unbalanced batches during optimization, rather than applying the weights as a multiplicative factor during the calculation of the loss function, we instead re-weight the probability of each particular instance in the training data being sampled at each batch.

\subsubsection{Matching}

In addition to changing the sampling weights of each sample in the training set, the most straight-forward possible approach to handling confounding suggested in \cite{Rao2017PredictiveConfounds} is matching the base rate across subgroups in the training data. The drawbacks to this approach are that it either requires deliberately collecting data that is balanced across subgroups in advance, or throwing out data. Since we could not go back and alter the data collection process for our dataset, in order to match the base rate of pneumonia in AP and PA radiographs in the training data, we had to delete 77,117 AP radiographs from the training data. This represented a substantial portion of the total data, amounting to 40\% of the samples negative for pneumonia in the CheXpert dataset, and 35\% of all samples in the training data.

\subsubsection{Include nuisance covariate in regression} Another potential approach to handle confounding suggested in \cite{Rao2017PredictiveConfounds} is to ``regress out" the effect of view on the outcome. We make use of the fact that the classification head of the DenseNet-121 is a logistic regression with the learned features (nodes of the last hidden layer, $H^{n-1}$) as covariates. Therefore, we simply append an extra feature for our covariate $V$ to the last layer $H^{appended} = [H^{n-1}_0, H^{n-1}_1, \cdots, H^{n-1}_i, V] $. We can then model the data using a standard logistic regression:

\begin{equation}
Y = \sigma(H^{appended}w + \beta),
\end{equation}

where $w \in \mathbb{R}^{h + | V |}$ is the vector of weights of the classification head, $\beta \in \mathbb{R}$ is the bias term for the classification head, and $\sigma(t) = \frac{1}{1 + e^{-t}}$.

We then train the modified DenseNet-121 following the exact same procedure as described in \autoref{sec:StandardTraining}. When evaluating our model in the external target domain, we remove the effect of the confounding variable by setting it equal to the mean across all samples.

\section{Results}

\begin{figure}
\centering
\includegraphics[width=0.85\linewidth]{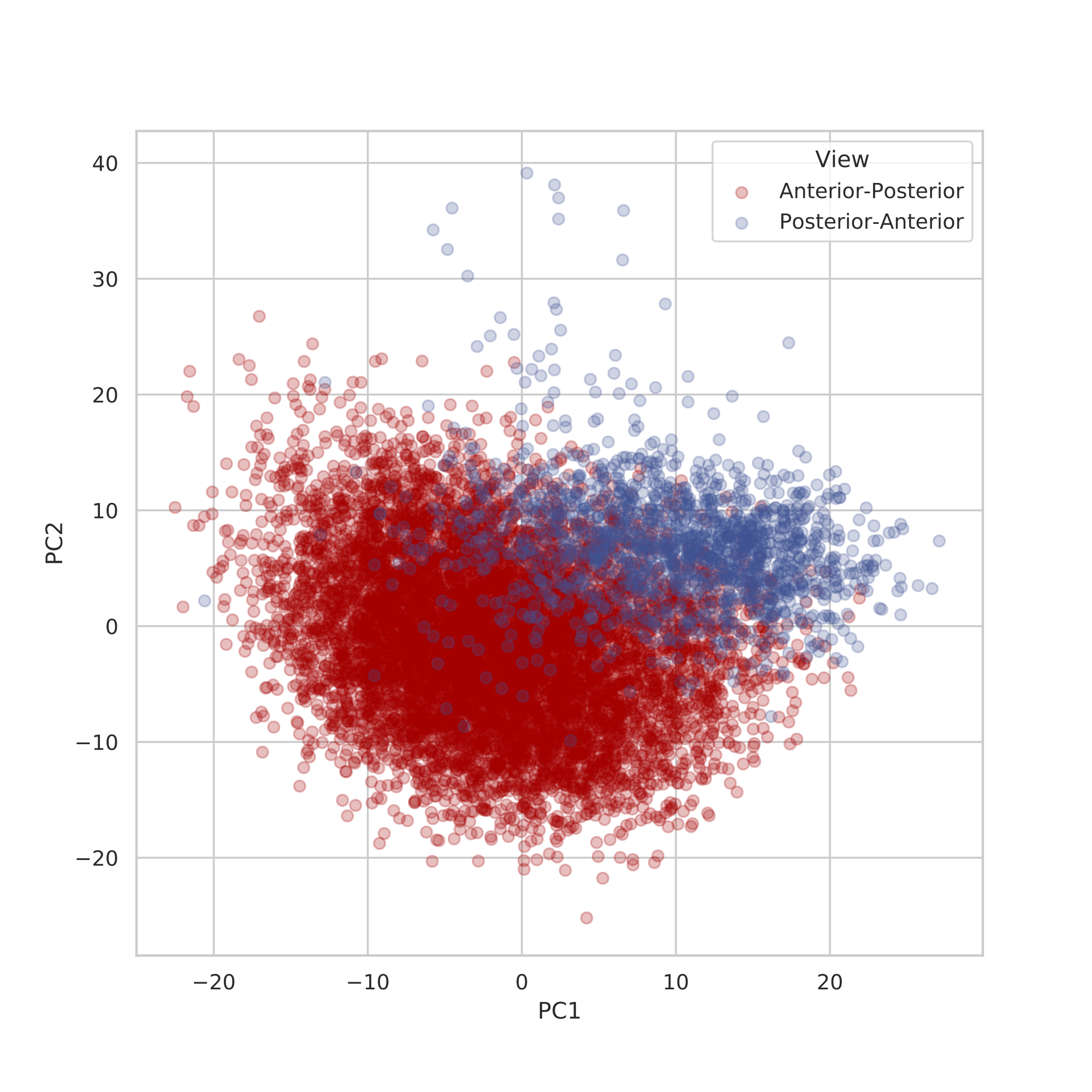}

\caption{A pretrained CNN with no task-specific supervision represents radiographs in a manner easily separable by view.}
\label{fig:ViewPCA}
\end{figure}

\begin{figure*}
\centering
\includegraphics[width=0.85\textwidth]{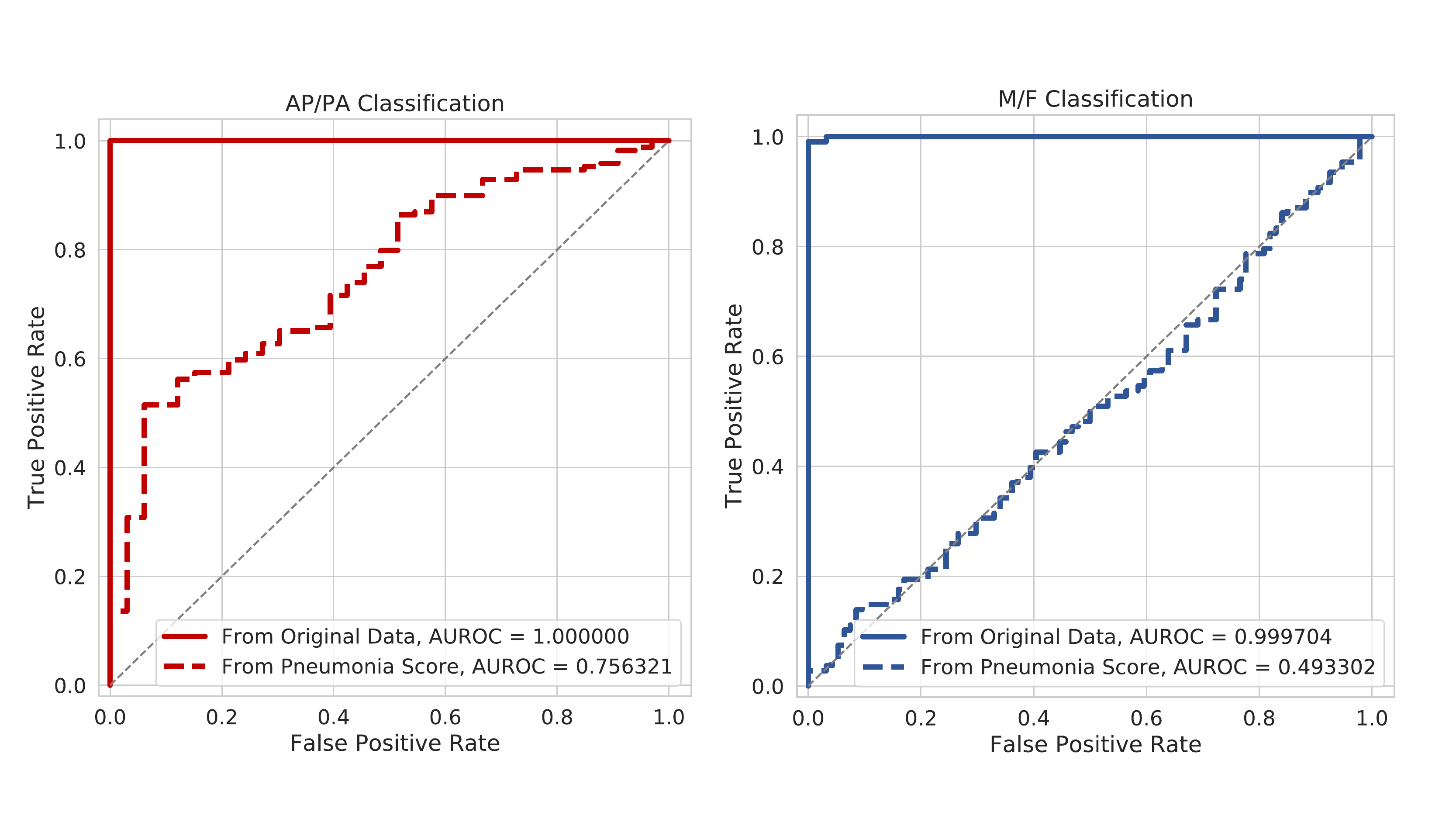}

\caption{The DenseNet-121 model architecture (solid lines) is capable of near-perfect prediction of the potential nuisance variables radiograph view and patient sex from the original image data. After training a DenseNet-121 to predict pneumonia from the original image data, we see that a simple feed-forward classifier is capable of predicting radiograph view using only the scalar-valued score output by the pneumonia model as input (dashed line, left). However, a neural network classifier fails to attain better than random performance at predicting patient sex from the same scalar-valued score (dashed line, right). This indicates that the pneumonia classification score is independent of patient sex, but not of radiograph view.}
\label{fig:FromDataScore}
\end{figure*}

\subsection{CNN pneumonia classifiers fail to generalize to external health datasets}
\label{sec:FailToGeneralize}

\begin{table}
  \caption{Pneumonia classifier performance (AUROC $\pm$ st. dev.) on held out test data from CheXpert dataset (Source), and held out test data from the external MIMIC dataset (Target). Standard deviations reported across three independent re-initializations of the training procedure. Best performance on external test data highlighted in bold and red.}
  \label{tab:generalization_performance}
  \begin{tabular}{cccl}
    \toprule
    Method & Source (Internal) & Target (External) \\
    \midrule
    Standard & 0.791 $\pm$ 0.016 & 0.703 $\pm$ 0.016 \\
    \color{red}{\textbf{Adversarial (Ours)}} & 0.747 $\pm$ 0.013 & \color{red}{\textbf{0.739 $\pm$ 0.001 }} \\
    Instance Weighting & 0.685 $\pm$ 0.049 & 0.648 $\pm$ 0.038 \\
    Covariate & 0.793 $\pm$ 0.008 & 0.715 $\pm$ 0.016 \\
    Matching & 0.684 $\pm$ 0.036 & 0.689 $\pm$ 0.024 \\
    \bottomrule
  \end{tabular}
\end{table}

To assess the generalization performance of standard deep learning approaches to pneumonia classification, we trained a classifier using the procedure described in \autoref{sec:StandardTraining} on data from the Stanford CheXpert dataset, then evaluated the model on both held-out patients from the same dataset (source performance) and held-out patients from the external MIMIC dataset (target performance). We evaluated performance using area under the ROC curve (AUROC), which evaluates the true positive rate and false positive rate attainable by the model across all possible thresholds.

We found that this model was able to achieve an AUROC of pneumonia classification of $0.791 \pm 0.016$ (see \autoref{tab:generalization_performance}). When we tested this same model on data from the PhysioNet MIMIC dataset, we found a substantial drop in performance, with the model only able to achieve an AUROC for pneumonia classification of $0.703 \pm 0.016$ (see \autoref{tab:generalization_performance}). This result again confirms the concerns raised in \cite{Pooch2019CanClassification} and \cite{Zech2018}, that state-of-the-art training and model architectures for deep learning medical imaging classifiers lead to models that do not generalize well to external datasets.

\subsection{Adversarial predictions improve model interpretability by identifying potentially confounded models}

\begin{figure*}
\centering
\includegraphics[width=0.85\textwidth]{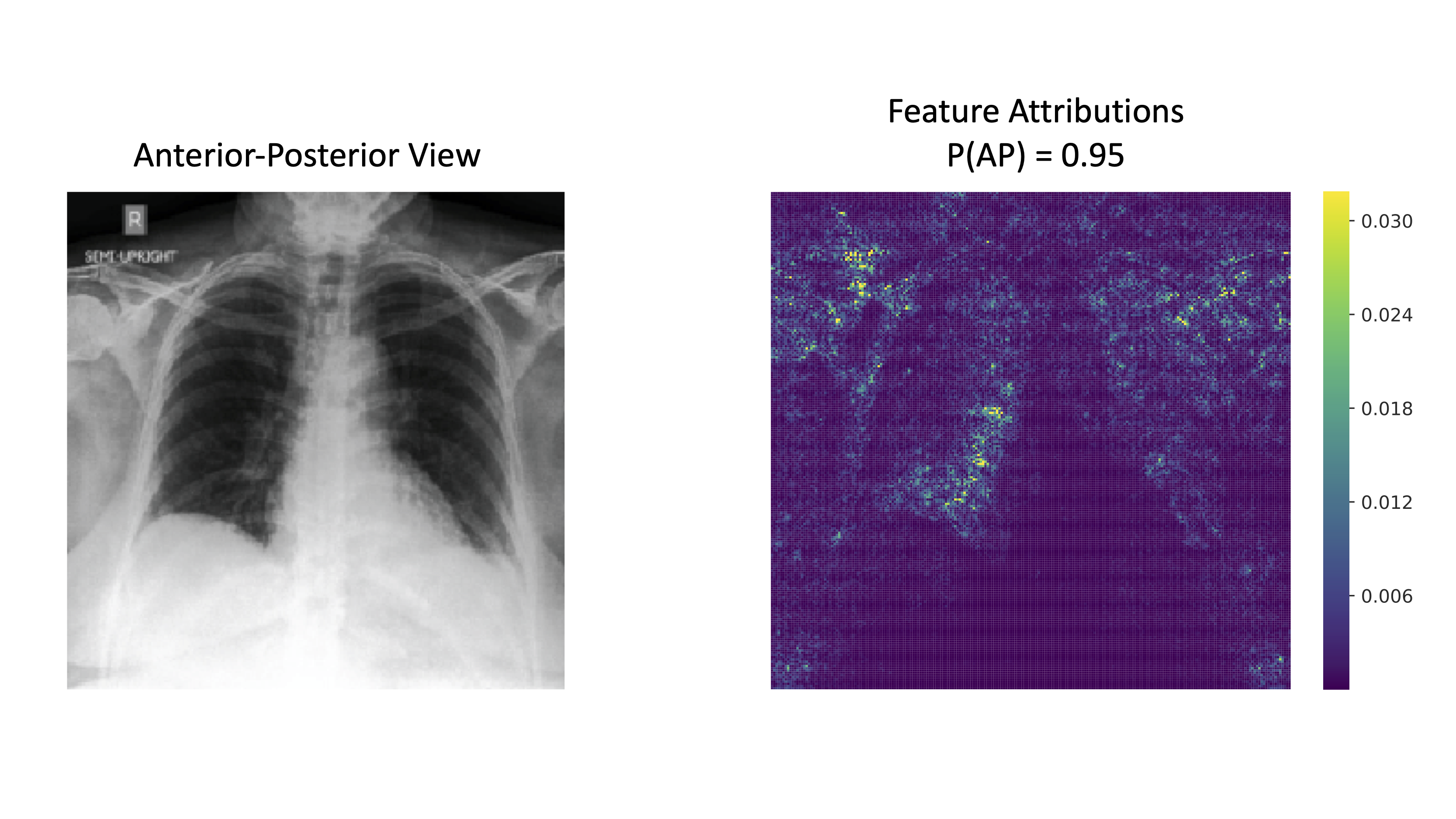}

\caption{Expected Gradients feature attributions for a DenseNet-121 classifier trained to predict radiograph view position (AP vs. PA). We see that while parts of the image like the laterality markers are important (and have previously been shown to be important confounders for identifying source hospital from chest radiographs \cite{Zech2018}), the most important pixels for identifying confounders are spread throughout the entire image, \emph{including} within the lung fields.}
\label{fig:APPAAttribs}
\end{figure*}

In this section we first show that state-of-the-art convolutional neural network (CNN) architectures are capable of detecting potential confounders given only pixel-level data. We then show how current approaches to model interpretability are of limited usefulness in determining whether or not a particular trained model depends on a potential confounder. We finally propose an approach based on training a neural network to predict the confounder from the model output, and show that it does a better job of identifying potential confounding.

\subsubsection{Pretrained networks separate radiographs on basis of view and sex without any supervision}
\label{sec:unsupervised}

To assess how easily CNNs separate radiographs on the basis of features other than pathology, we examined the features extracted by a DenseNet-121 pretrained on ImageNet before \emph{any} training on chest radiographs (\autoref{fig:ViewPCA}). We randomly sampled 10,000 radiographs and applied the DenseNet-121 features submodule to them (i.e. the entire model except the classification head). We then average pooled over the last two dimensions to get 1024 features for each sample. To visualize how different sorts of radiographs were spread over these pretrained features, we performed principal components analysis on the resulting matrix, and compared the distributions of different subsets of the data along the principal components. We found that the ImageNet-pretrained DenseNet-121 easily separates chest radiographs on the basis of their view, as AP and PA radiographs are embedded in different parts of the last layer.

\subsubsection{Both pretrained and scratch-trained networks exploit confounders}
Since pretraining alone was so easily able to separate confounders, we wanted to rule out the possibility that pretraining on ImageNet was contributing to the confounding or poor generalizability we observed in \autoref{sec:FailToGeneralize}, especially in light of recent results on the potentially limited benefit of transfer learning for medical imaging \cite{Raghu2019Transfusion:Imaging}. Therefore, we also tried training a deep CNN architecture from randomly initialized weights using the same training approach and same CheXpert data (see \autoref{sec:ResNet}). While we found that while this model was able to achieve comparable classification performance on held-out test data from the CheXpert dataset, it generalized significantly worse to the external target domain MIMIC data, indicating that ImageNet pretraining may actually be helpful for model robustness. We therefore are able to conclude that not only do CNNs have access to potential confounders using only pixel-level data, but that this problem can not be solved just by removing pretraining as a training step.

\subsubsection{CNNs can detect potential confounders from image data with high accuracy}

\begin{figure*}
\centering
\includegraphics[width=0.85\textwidth]{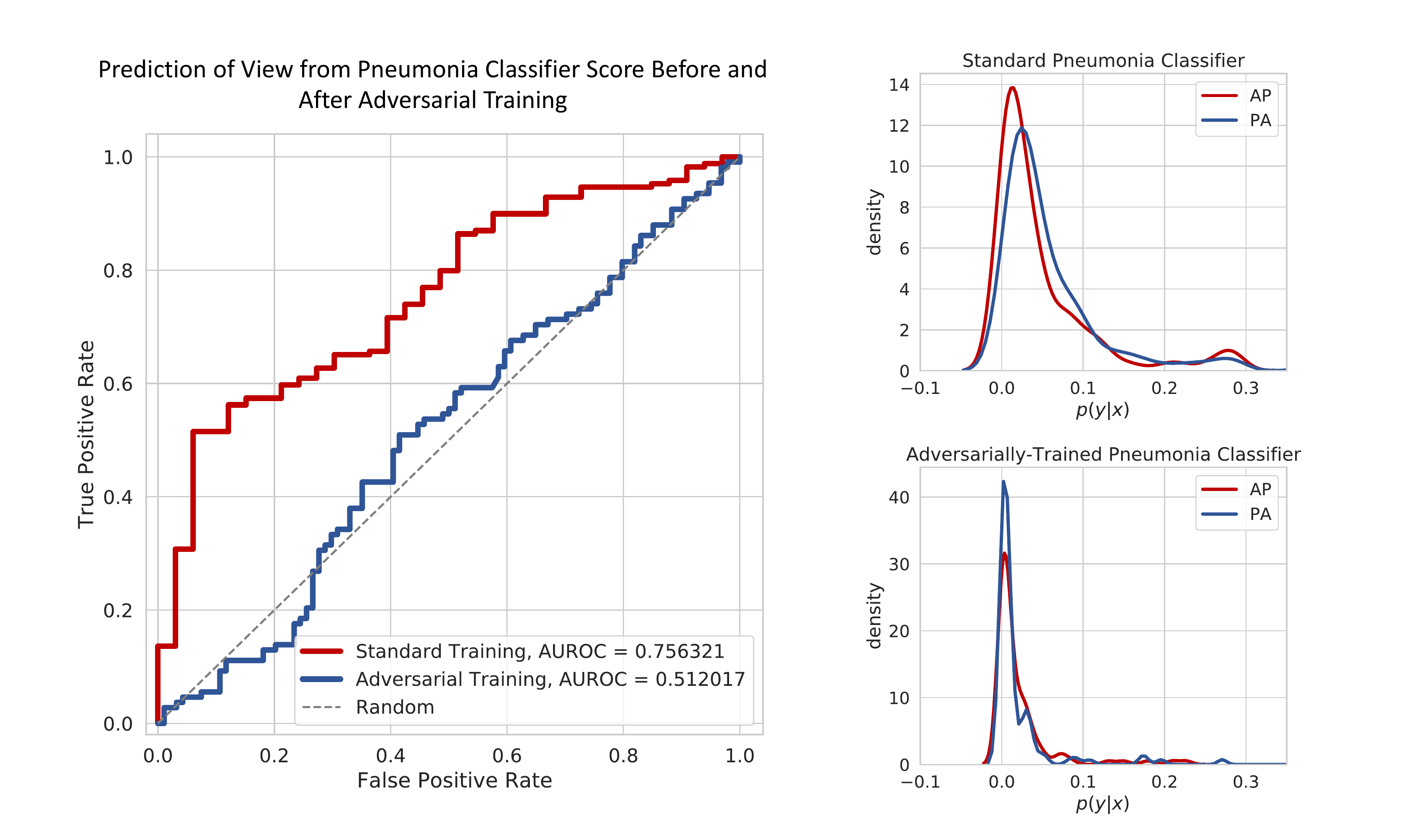}

\caption{Adversarial training learns a pneumonia score that is independent of view. LEFT: ROC curves for the prediction of radiograph view (AP vs. PA) from a classifier's pneumonia score for a classifier trained with a standard approach (red) and for a classifier trained with our adversarial approach (blue). View can be predicted with relatively high accuracy just using the pneumonia score from the standard classifier, indicating that this model's output and view are not independent. After adversarial training, view can no longer be predicted with better-than-random accuracy, indicating that the output of this classifier is independent of view. RIGHT: When we look at the distribution of pneumonia scores actually output by the two models (Top: Standard, Bottom: Adversarial), we see that the distributions are not identical between AP and PA subgroups in the standard training model, but are much more closely matched betweeen the AP and PA subgroups in the adversarially-trained model.}
\label{fig:IndependentScore}
\end{figure*}

A previously proposed approach for detecting potential confounders has been to evaluate how well that confounder can be predicted from the original data \cite{Zech2018}. When we train the same architecture CNN using the same training procedure to predict nuisance variables like sex or radiographic view from the chest radiograph data, we find that our models are capable of predicting these variables with incredibly high accuracy (\autoref{fig:FromDataScore}). For example, we see that a model can predict view with an AUROC of 1.0, perfectly classifying every example from the held out test data. Similarly, we see that this same model architecture is capable of predicting patient sex with an AUROC of 0.9997, again on held out test data. This result establishes that even if potentially confounding nuisance variables like radiograph view or patient sex are not explicitly included in the input features of CNN classifiers, deep CNN architectures are able to extract them with high accuracy from the pixel-level features of the radiographs, allowing them to still be used in classification.

While this result indicates that CNNs can detect potential confounders from just the radiograph data, it does give us any way to tell whether or not a particular model is invariant to a particular confounder. For example, in the CheXpert dataset, the base rate of pneumonia in male patients is $2.39\%$ while the base rate of pneumonia in female patients is $2.42\%$. Therefore, even though we have seen that a CNN \emph{can} identify whether a radiograph is from a male or female patient with high accuracy, it seems likely that a model would already be invariant to a feature that does not have an association with the outcome of interest.

Saliency maps are another previously proposed approach for understanding model behavior \cite{Sundararajan2017AxiomaticNetworks,Smilkov2017Smoothgrad:Noise,Selvaraju2016Grad-CAM:Localization}. These methods highlight the pixels or regions that were most important for the classifier in a given image. We therefore used Expected Gradients, a pixel-level feature attribution method \cite{erion2019learning}, to generate saliency maps to help understand which pixels were important for classifying view from radiographs (see \autoref{fig:APPAAttribs} and Appendix section \autoref{sec:FeatureAttributions} for more details). We observe that there is no specific region in the image that is indicative of PA vs. AP view. While both the laterality marker and text marker on the image are important for classification of view, pixels throughout the entire image, including within the lung fields, are also important for this prediction. Therefore, saliency map-based approaches are also not necessarily useful for identifying whether a model is invariant to a confounder or not.

\subsubsection{Confounders can be detected directly from score}

While seeing if nuisance variables can be predicted from the images can help understand if a confounded model \emph{could} be learned from some data, it does not help identify how much a \emph{particular model} is actually invariant to confounders. To assess this, we instead evaluate how well a neural network model (adversary) can classify the confounder of interest using only the scalar output score of the model we care about as input. In our case, this quantifies the dependence between the output score for pneumonia $S$ and the confounding variable $V$ by measuring the difference between the two distributions $p(S|X, V = \text{AP})$ and $p(S|X, V = \text{PA})$, and is well-justified as an empirical approximation to the $\mathcal{H}$-divergence discussed in \cite{Ganin2016Domain-adversarialNetworks,Edwards2015CensoringAdversary}. For a classifier where the output with respect to our class of interest, $S$ is independent of $V$, prediction of $V$ from $S$ should be random, while $S$ not independent of $V$ will lead to better than random prediction. As an adversary, we trained a simple feed-forward network with 3 hidden layers of 32 nodes.

While both view and sex were nearly perfectly classified from the original data, when we first train a DenseNet-121 classifier to predict pneumonia from chest radiographs, then try to predict view and sex from the predicted probability of pneumonia, we see that our model attains far greater performance at predicting radiograph view than sex, and that patient sex is not predicted better than random (\autoref{fig:FromDataScore}). Therefore, we can conclude that while the pneumonia classifier is likely independent of sex, it is potentially not invariant to view.

\subsection{Adversarial training can increase model robustness by controlling for confounders}

\subsubsection{Adversarial framework learns view-independent classifier}

\begin{figure*}
\centering
\includegraphics[width=0.85\textwidth]{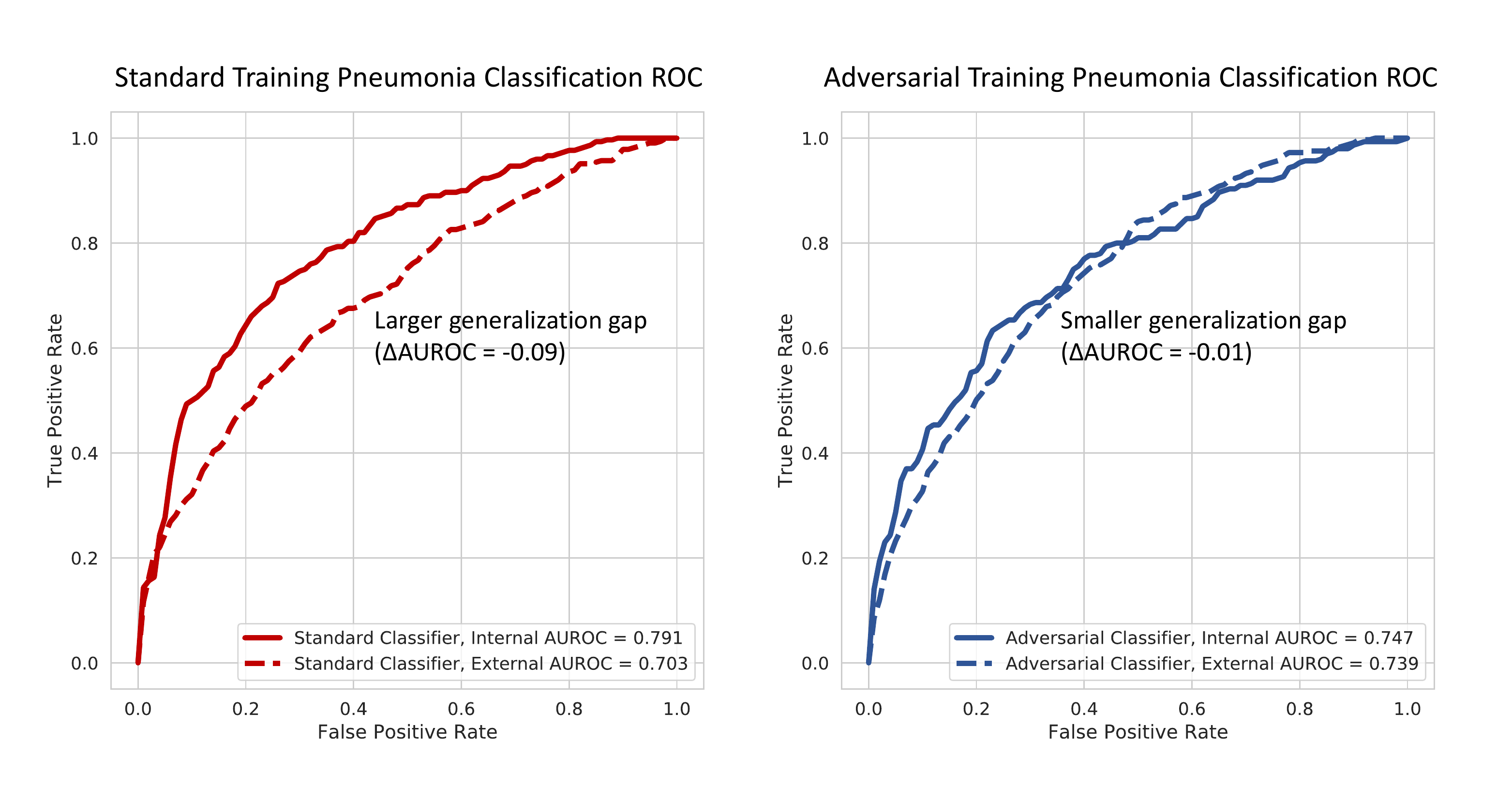}

\caption{Adversarial training leads to less performance drop and significantly better performance when classifier is tested on data from a hospital system external to the one the training data comes from.}
\label{fig:GeneralizationGapAUROC}
\end{figure*}

Following the insight of the previous section, we can directly optimize for a classifier that learns a score for our class of interest $S$ that is independent of view using an adversarial framework. Prior to adversarial training, an adversary neural network could predict the counfounder with relatively high accuracy given only the score (\autoref{fig:FromDataScore}). Following our adversarial optimization procedure (\autoref{sec:AdvTraining}), a neural network is not able to predict the confounder any better than random accuracy (see \autoref{fig:IndependentScore}, left). Furthermore, when we look at the actual score distributions output by our model, we find that they are more closely matched within the two different view subgroups (see \autoref{fig:IndependentScore}, right top and right bottom). While we mainly present results for the binary view variable, one strength of our approach is that it can be applied to any sort of nuisance variable, including continuous-valued variables like age (see \autoref{sec:AgeAppendix}).

We also find that looking at the predictive performance of the adversarial classifier is far more indicative of model behavior than saliency map-based approaches in this case. When we plot saliency maps (see \autoref{fig:EGAttribs} in the Appendix, as described in \autoref{sec:FeatureAttributions}) we can see that there are definite differences in the pixel-level attributions. Furthermore, it appears that the important pixels are more localized to the lung fields in the adversarially-trained model than in the standard model. However, it is difficult to quantitatively assess to what extent that is the case, and since we have shown that pixels throughout the entire image are important for view classification by CNNs, it is very difficult to answer whether or not a model is confounded by view or not based only on its pixel-level feature attributions.

\subsubsection{View-independent classifier generalizes better to unseen target domain}

In addition to being able to learn a classifier that is independent of view, we find that adversarial training also is able to learn a model that generalizes better to external target domain test data (see \autoref{tab:generalization_performance}). When we compare the performance of the adversarial model to the standard model, we find that while the adversarial model attains slightly worse performance on the source domain  (AUROC $= 0.747 \pm 0.013$ vs. AUROC $= 0.791 \pm 0.016$), it attains better performance on the target domain (AUROC $= 0.739 \pm 0.001$ vs. AUROC $= 0.703 \pm 0.016$).

When we compare to the other baseline methods for controlling for confounding (instance weighting, including the covariate, and matching), we find that adversarial training also outperforms these methods. Of these other methods, we find that including the confounder as an additional covariate in modeling is the most effective, followed by matching, then the instance weighting resampling scheme.

\begin{figure*}
\centering
\includegraphics[width=0.85\textwidth]{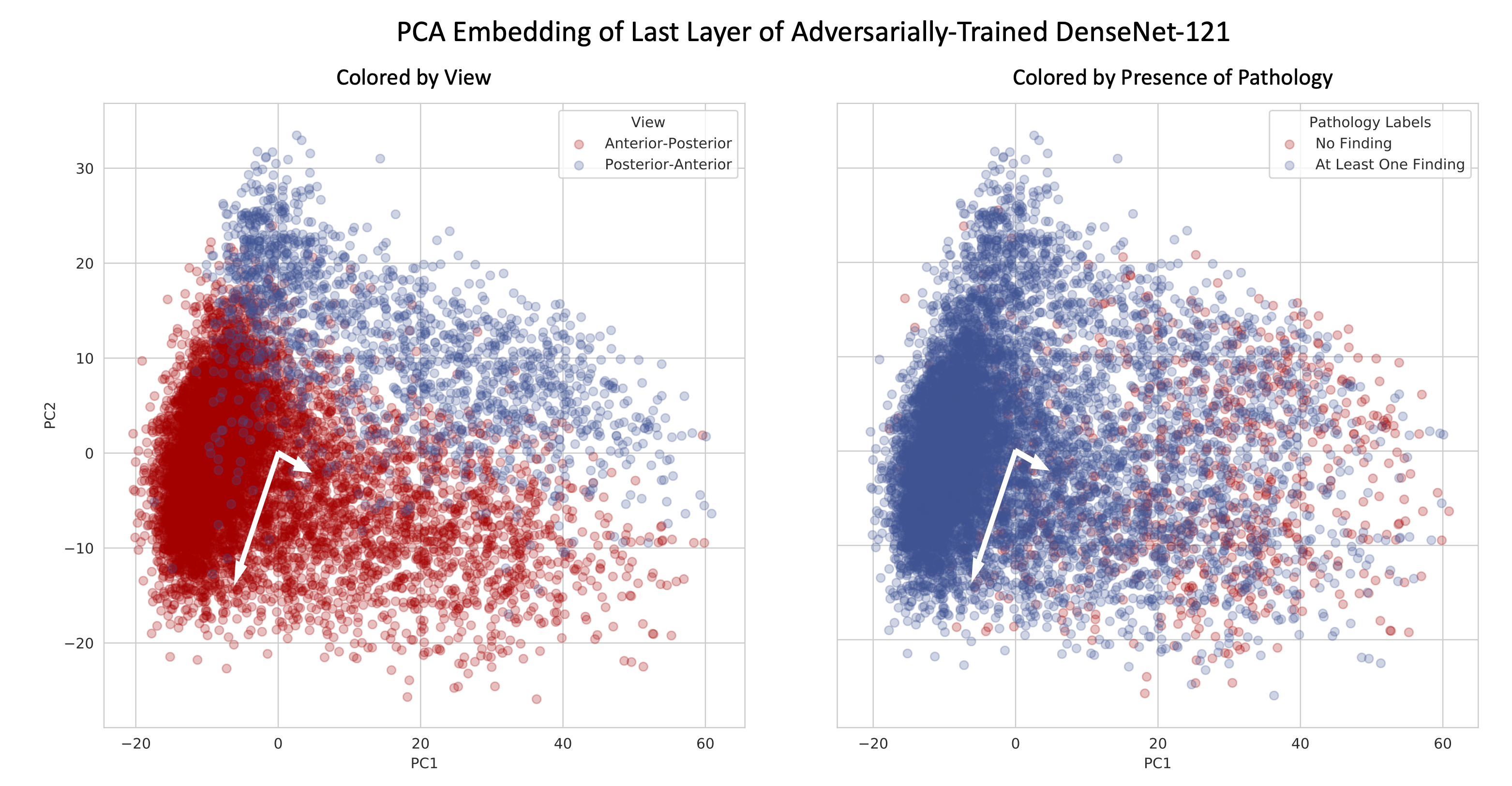}

\caption{Adversarial training leads to a final representation where general pathology is orthogonal to view. White arrows indicate magnitude and direction of view and pathology classification weight vectors.}
\label{fig:SupervisedEmbedding}
\end{figure*}

\subsubsection{Adversarial training learns a representation where pathology is independent of view}

While our adversarial approach only explicitly constrains the final output score to be independent of the nuisance variable $V$ representing view, we wanted to see how the earlier representations in the DenseNet-121 were impacted by this approach. We therefore take the output of the last dense layer before the classification head and average pool over the last two dimensions, and then perform principal components analysis in the same way as we did for the ``unsupervised'' ImageNet-pretrained classifier in section \autoref{sec:unsupervised}.

The representation in the last layer of our adversarially-trained classifier is interesting, in that it is able to learn an embedding where the axis of differentiation separating the two views seems to be orthogonal to the axis representing pathology (see \autoref{fig:SupervisedEmbedding}). When we plot the first two principal components of the radiograph embedding and color by view, we see that the views are separated from the bottom left of the plot towards the top right. When we color the same embedding by pathology, the images with no findings are separated to the bottom right, while images containing pathology separate to the top left. 

To quantify if this adversarially-trained representation has a more orthogonal relationship between view and pathology than a classifier with standard training, we learned a simple logisitic regression classifier using the first two principal components of the last layer embeddings of the standard and adversarially trained classifiers as input. The output for prediction was either view or pathology. We then measured the linear correlation ($r$) between the weight vectors of the two linear classifiers.

We found over a 10-fold decrease in the correlation between the view and pathology vectors in the final embeddings from the standard to adversarially-trained models (decreasing from $r = 0.1974$ to $r = 0.008$), indicating that the view-axis was substantially more orthogonal to the pathology-axis in the adversarially-trained classifier. Again, this was particularly remarkable in that there was no constraint to learn a more independent representation in the hidden layers of the model.

\section{Discussion}

Our results demonstrate that an approach based on adversarial optimization is capable of learning more robust medical imaging classifiers. For the specific case of chest radiographs, we show that a pneumonia classifier trained to be independent of view is more stable to dataset shift, attaining better generalization performance when tested on radiographs from an external dataset. Finally, our results show that attempting to predict potential nuisance variables directly from a model's output score can be a valuable tool for model interpretability, indicating whether or not a particular model is independent of potential confounders. While any measure of the difference in the distributions of the model's output conditional on potential confounders is likely to work well, we believe that our approach is well-suited in that it also lends itself naturally to a technique to create confounder-invariant models.

Examination of the causal diagram relating chest radiographs to pneumonia points to important future research directions. Our experiments showed increased stability to dataset shift at the expense of decreased performance on new samples from the same hospital system as the training data. Given the causal diagram, where view mediates the relationship between the presence of pneumonia and the pixel features of the chest radiograph, it is not surprising that controlling for view should decrease performance. We note, however, that pneumonia is a diagnosis that is made in the context of clinical evidence of disease, and a disease where there is not necessarily perfect concordance between severity of symptoms and radiographic evidence of infiltrate \cite{vanVugt2013DiagnosingRadiography,Niederman1993GuidelinesTherapy,Oakden-Rayner2018CheXNet:Review,Botz2017AMeasured}. In the description of the creation of the ``Pneumonia'' label in the CheXpert dataset, the authors note that while pneumonia is a clinical diagnosis, ``Pneumonia... was included as a label in order to represent the images that suggested primary infection as the diagnosis,'' suggesting that clinical information may play a role in labeling \cite{Irvin2019Chexpert:Comparison}. Disentangling the relationship between radiographic evidence of consolidation, the clinical presence of pneumonia symptoms, and the influence of the latter on the labeling of the former in these datasets could be helpful.

Finally, while we showed results from controlling radiograph view (and patient age), we expect that future work could show even more benefits from applying our approach to a wider variety of variables, both individually and in combination. However, it would be required for these variables to be recorded as metadata in datasets. As more and more additional variables are recorded in medical imaging datasets, and the causal relationships between these variables are better explicated, we expect the potential benefit of our approach to further increase.

\begin{acks}
We would like to thank Samantha Gilbert, Pascal Sturmfels, Hugh Chen, Nicasia Beebe-Wang and Alex Okeson for their feedback on the manuscript. We would also like to thank all of the members of Prof. Su-In Lee's lab for their valuable general feedback on the project.
\end{acks}

\bibliographystyle{ACM-Reference-Format}
\bibliography{references}

\appendix

\section{Subgroup base rate imbalance and generalization performance}
\label{sec:BaseRateAlteration}

\begin{table}
  \caption{Comparing extent of base rate imbalance between AP and PA in training data and generalization performance of pneumonia classification on external MIMIC test data}
  \label{tab:pneumo_base_rate}
  \begin{tabular}{cc}
    \toprule
    Base Rate Imbalance & MIMIC Pneumonia AUROC \\
    \midrule
    1-to-1 & 0.6508 \\
    2-to-1 & 0.6237 \\
    10-to-1 & 0.5847 \\
    100-to-1 & 0.5749 \\
    \bottomrule
  \end{tabular}
\end{table}

Since radiograph view labels were not provided in the MIMIC dataset, we wanted to ensure that the difference in the base rate of pneumonia between view subgroups in the CheXpert training data was really an important factor contributing to poor generalization performance. Inspired by the ``engineered relative risk experiment'' in \cite{Zech2018}, we therefore created four synthetic subsamples of the CheXpert dataset with differing base rates of pneumonia between AP and PA radiographs. The first had a balanced ratio, the second had a 2-to-1 imbalanced ratio, the third had a 10-to-1 imabalanced ratio, and the fourth had a 100-to-1 imbalanced ratio . Each dataset contained 20,000 images total (10,000 from each AP and PA), and the base rate of pneumonia \emph{overall} was held constant at $5\%$. This means that, for example, in the 100-to-1 imbalanced setting there were 10 AP radiographs that were positive for pneumonia, 9,990 AP radiographs that were negative for pneumonia, $990$ PA radiographs that were positive for pneumonia, and 9,010 PA radiographs that were negative for pneumonia. It was necessary to decrease the sample number in the synthetic datasets to be able to achieve greater ratios of base rate imbalance between AP and PA subgroups.

We trained models on the synthetic datasets using the standard training procedure, tested on the external MIMIC dataset, then measured how much of the predictive performance was lost as the base rate became more imbalanced. In the balanced synthetic dataset, we had a baseline AUROC of 0.6508 (see \autoref{tab:pneumo_base_rate}. As we increased the base rate difference to 2-to-1, we saw a significant drop in predictive performance (AUROC of 0.6237). As we continued to increase the base rate difference to 10-to-1 and 100-to-1 we found that the predictive performance continued to drop. We therefore were able to conclude that as the base rate difference between AP and PA radiographs was exacerbated, the generalization performance of our classifiers was decreased.

\section{Randomly initialized network}
\label{sec:ResNet}
To rule out the possibility that the reason these models were using confounding information and generalizing poorly was due to the initialization with weights that were pretrained on ImageNet, we also tried training deep CNNs with weights that had been randomly initialized. In addition to a DenseNet-121, we also tried training a ResNet-50 architecture \cite{He2016DeepRecognition}. Other than the change in initialization, the training procedure was identical to that described in the standard training section for both architectures.

While the source domain performance (area under ROC on CheXpert data) for the ResNet was comparably high to the performance attained by the ImageNet-pretrained DenseNet-121, the performance gap when testing on external target domain test data (area under ROC on MIMIC data) was actually much more significant, representing a drop in AUROC of 0.18 (see \autoref{tab:random_class}). Furthermore, for the DenseNet architecture, both source and target domain performance decreased.

\begin{table}
  \caption{Randomly Initialized Classifier Performance (AUROC) on Source Domain (CheXpert) and Target Domain (MIMIC)}
  \label{tab:random_class}
  \begin{tabular}{cccl}
    \toprule
    Method & Source (Internal) & Target (External) & $\Delta$\\
    \midrule
    ResNet-50 & 0.7829 & 0.5992 & -0.18\\
    DenseNet-121 & 0.7098 & 0.5674 & -0.14\\ 
    \bottomrule
  \end{tabular}
\end{table}

\section{Feature attributions}
\label{sec:FeatureAttributions}

To generate saliency maps, we used the recently developed state-of-the-art method for feature attributions known as Expected Gradients \cite{erion2019learning}. Expected Gradients is an extension of the Integrated Gradients feature attribution method \cite{Sundararajan2017AxiomaticNetworks}. For a model $f$, the \textit{integrated gradients} value for feature $i$ is defined as: 
\[
    \textrm{IntegratedGradients}_i(x, x') = (x_i - x_i') \times \int_{\alpha = 0}^1 \frac{\delta f(x' + \alpha \times(x - x'))}{\delta x_i} \delta \alpha, 
\]
where $x$ is the target input and $x'$ is baseline input. In medical imaging models, it is not clear what image would serve as a reasonable baseline. Therefore, we use the \textit{expected gradients} value for feature $i$, which does not need a background reference and uses the entire training data in expectation:

\[
    \textrm{ExpectedGradients}_i(x) =  \mathop{\mathbb{E}}_{x' \sim D, \alpha \sim U(0, 1)} \bigg [ (x_i - x_i') \frac{\delta f(x' + \alpha \times(x - x'))}{\delta x_i} \bigg ].
\].

To visualize pixel-level feature importances, we plot a heatmap where the color intensity encodes the magnitude of each attribution averaged across the three image channels. We clipped attributions in magnitude at the $99.9$th percentile for the visualization.

\begin{figure*}
\centering
\includegraphics[width=0.55\textwidth]{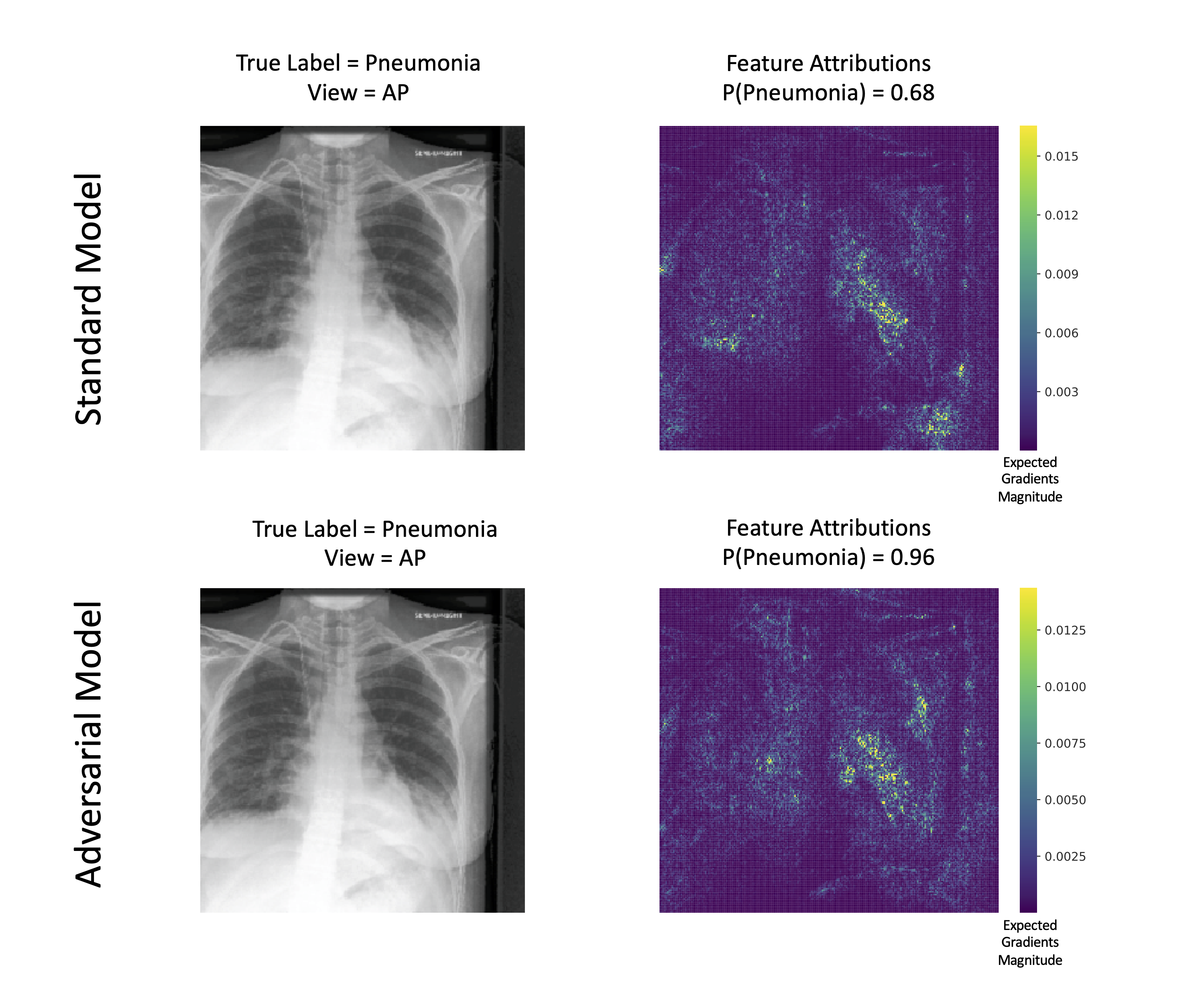}

\caption{Differences in Expected Gradients attributions for standard vs. adversarial classifiers. We notice that the adversarial model places less importance on the pixels in the lower right hand corner of the image outside of the lung fields compared to the standard model. We also observe that it is difficult to quantify from pixel-level attributions how important the view confounder was to the model. This is a major limitation of saliency map approaches. While they can help rule-in the possibility of questionable model behavior, it is difficult to rule-out the possibility of undesirable confounding.}
\label{fig:EGAttribs}
\end{figure*}

\section{Subgroup base rate imbalance and adversarial confounding score}
\label{sec:CounfoundednessDetection}

In the models trained on the synthetic datasets in \autoref{sec:BaseRateAlteration}, we saw that as the relationship between view and pneumonia became more confounded, the models performed less well on external test data. We wanted to ensure that the predictive performance of an adversary trained to predict view from the output of these models increased as the confounding was exacerbated. We therefore trained an adversary to predict view given the output of these three models, and found that as the confounding was exacerbated, the predictive performance of the adversary increased (see \autoref{tab:adversary_base_rate}, which also includes a comparison with the adversarial performance achieved on the standard data which has a 2-to-1 imbalance).

\begin{table}
  \caption{Comparing extent of base rate imbalance between AP and PA in training data and performance on adversarial neural network trained to predict view from output score of classifier.}
  \label{tab:adversary_base_rate}
  \begin{tabular}{cc}
    \toprule
    Base Rate Imbalance & Adversarial AUROC \\
    \midrule
    1-to-1 & 0.4956 \\
    2-to-1 & 0.7563 \\
    10-to-1 & 0.9180 \\
    100-to-1 & 0.9436 \\
    \bottomrule
  \end{tabular}
\end{table}

\section{Age distribution results}
\label{sec:AgeAppendix}

One major advantage of the adversarial approach to deconfounding when compared to matching or balancing based approaches is that it is remarkably easy to control for continuous variables. The architecture of the adversarial neural network is changed by simply replacing the sigmoid activation function at the output with a linear activation, and replacing the binary cross-entropy loss function for the adversary network with a mean squared error loss function. Then training proceeds identically to the procedure described in \autoref{sec:AdvTraining}. To demonstrate that this approach works for continuous variables, we show results for controlling the only continuous potential nusiance variable present in the CheXpert dataset, patient age (see \autoref{sec:AgeAppendix}). 

After adversarial training a classifier to produce a pneumonia score independent of age, we wanted to evaluate how well our approach changed the age-conditional score distributions of our classifier. We split the data into four separate subgroups thresholded by age: patients younger than 45, patients with age between 45 and 65, patients between 65 and 85, and patients older than 85. We then measured all of the pairwise distances between these subgroups, in both the standard and adversarially trained classifiers. We found that in general, the score distributions were as similar or more similar between pairs of age subgroups in the adversarially-trained classifier than in the standard classifier. Since we do not expect age to be a meaningful confounder for pneumonia prediction that is likely to shift between hospital datasets, we would not expect controlling for age to improve predictive performance. We do not find any increased predictive performance in external target domain test data by controlling for age (AUROC of 0.695 after controlling for age as compared with an AUROC of 0.703 with standard training).

\begin{figure*}
\centering
\includegraphics[width=0.40\textwidth]{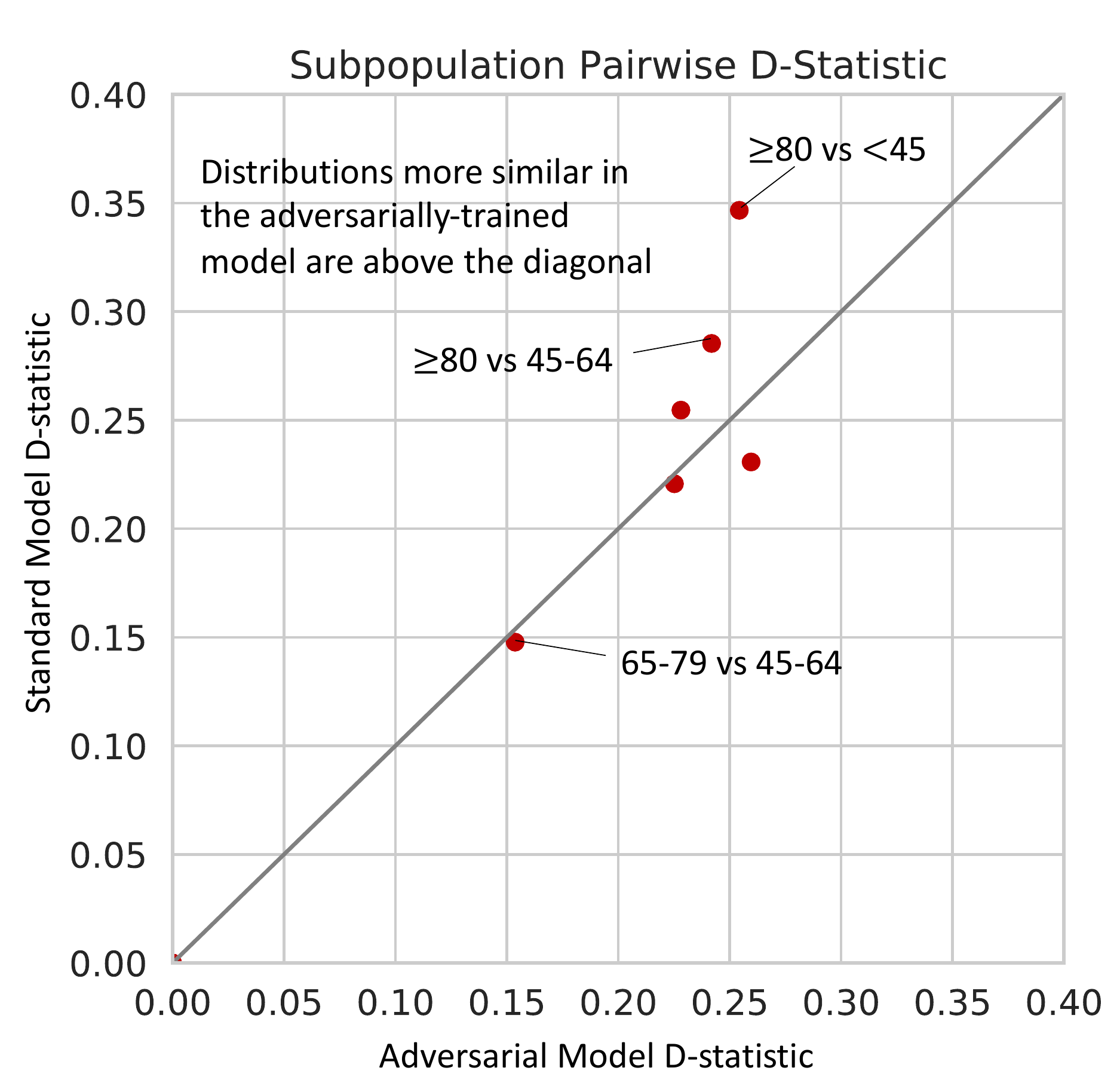}

\caption{Adversarial training can learn a score that is independent of continuous random variables. To evaluate how similar the distribution of model scores were across different age subgroups, we divided the samples into four age subgroups. Each dot in the plot above represents the distance between the score distributions of two of the age subgroups (as measured by KS D-statistic). The y-axis shows the D-statistic for the scores output by the standard classification model, while the x-axis shows the D-statistic for the scores output by the adversarially-trained classification model. Dots that are above the diagonal indicate subgroups whose distributions became more similar under the adversarially-trained model.}
\label{fig:AgeDist}
\end{figure*}

To measure the difference between the age-subgroup score distributions, we used the Kolmogorov-Smirnov D statistic, which is defined as

\begin{equation}
    D_{n,m} = \textrm{sup}_x | F_{1,n}(x) - F_{2,m}(x) |,
\end{equation}
where $F(x)$ is the empirical distribution function. This statistic takes values between $0$ and $1$, where values closer to $0$ indicate more similar distributions, while values closer to $1$ indicate less similar distributions. In the plot, we compare each of the six possible pairs of age subgroups on their similarity in the standard classifier and their similarity in the adversarially-trained classifier \autoref{fig:AgeDist}. For 3 out of the 6 pairs, we see a significant increase in score distribution similarity. For 1 pair we see a decrease, and for 2 pairs the score distribution similarity remains roughly constant. We notice that the distributions that are initially more divergent in the standard classifier are improved the most, and correspond to the greatest age difference.

\end{document}